# How Human Motion Prediction Quality Shapes Social Robot Navigation Performance in Constrained Spaces

**Andrew Stratton**
University of Michigan at Ann Arbor
Ann Arbor USA
arstr@umich.edu

**Phani Teja Singamaneni**
LAAS-CNRS - University of Toulouse
Toulouse France
Inria
Nancy France
phani-teja.singamaneni@inria.fr

**Pranav Goyal**
University of Michigan at Ann Arbor
Ann Arbor USA
prgoyal@umich.edu

**Rachid Alami**
LAAS-CNRS - University of Toulouse
Toulouse France
rachid.alami@laas.fr

**Christoforos Mavrogiannis**
University of Michigan at Ann Arbor
Ann Arbor USA
cmavro@umich.edu

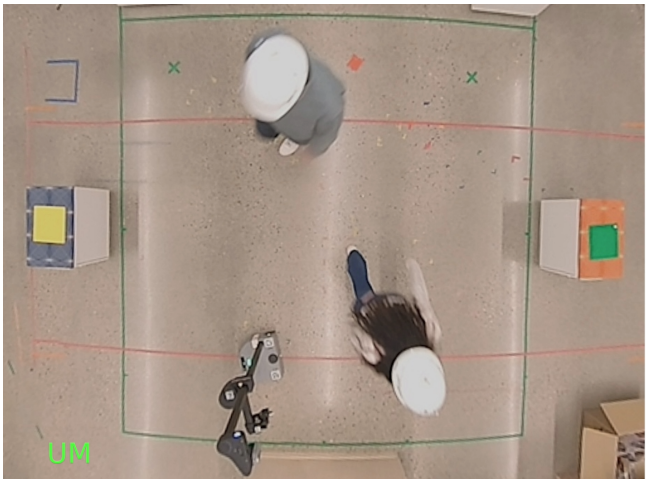 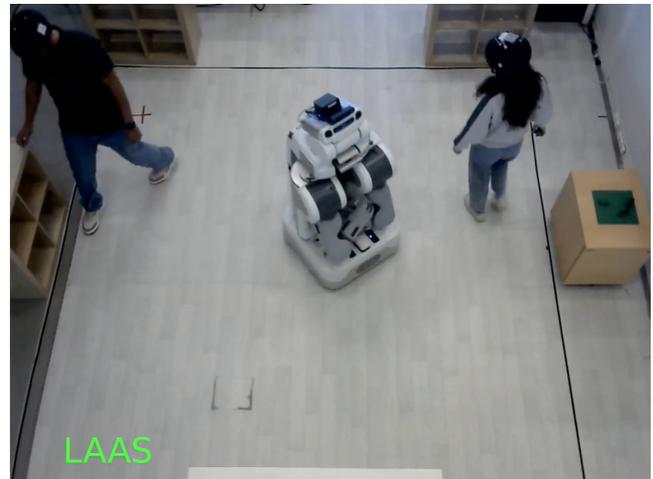

**Figure 1**: Still from our experiments. Two users and a robot navigate a constrained workspace. We mirrored the same study setup at UM and LAAS, using a Hello Robot Stretch (left), and a Willow Garage PR2 (right), collecting interactions with a total of $N = 80$ subjects. A video supplement with excerpts from our experiments can be found at https://youtu.be/QGd5vQMhXqY.

## Abstract

Motivated by the vision of integrating mobile robots closer to humans in warehouses, hospitals, manufacturing plants, and the home, we focus on robot navigation in dynamic and spatially constrained environments. Ensuring human safety, comfort, and efficiency in such settings requires that robots are endowed with a model of how humans move around them. Human motion prediction around robots is especially challenging due to the stochasticity of human behavior, differences in user preferences, and data scarcity. In this work, we perform a methodical investigation of the effects of human motion prediction quality on robot navigation performance, as well as human productivity and impressions. We design a scenario involving robot navigation among two human subjects in a constrained workspace and instantiate it in a user study ($N = 80$) involving two different robot platforms, conducted across two sites from different world regions. Key findings include evidence that:

1) the widely adopted average displacement error is not a reliable predictor of robot navigation performance and human impressions; 2) the common assumption of human cooperation breaks down in constrained environments, with users often not reciprocating robot cooperation, and causing performance degradations; 3) more efficient robot navigation often comes at the expense of human efficiency and comfort.

## CCS Concepts

• **Computer systems organization** → **Robotic autonomy**; *Robotic control*; • **Human-centered computing** → **Laboratory experiments**; *Collaborative interaction*; • **Computing methodologies** → *Cognitive robotics*; *Robotic planning*.

## Keywords

Human motion prediction, Social robot navigation, Benchmarking

## 1 Introduction

Mobile robots have the potential to revolutionize important sectors like healthcare, manufacturing, and fulfillment by efficiently tackling delivery and assembly tasks alongside humans. While mobile robots already appear prominently in warehouses (e.g., Kiva systems [7]), safety and efficiency concerns do not allow humans and robots to move freely in the same areas. Addressing these concerns around unstructured and dynamic human behavior motivates research on social robot navigation (SRN) [33, 55], which can enable an entirely new paradigm for human-robot co-working in shared spaces. Essential to the success of SRN algorithms are mechanisms of human motion prediction (HMP), which enable anticipation of human behavior. HMP modeling empowers SRN algorithms to negotiate shared spaces [35, 36], and helps proactively reduce the robot's intrusion of humans' personal and intimate proxemic zones [12], which may impact human comfort and workload [42]. Research in HMP [44] is primarily concerned with the task of offline prediction on human datasets, treating robot deployment as a downstream task. However, robot deployment introduces unique challenges like out-of-distribution instances (OOD), robot morphology artifacts, and novelty effects, all of which may also impact robot performance and human acceptance. These factors motivate the study of HMP and SRN as a coupled problem [33].

Algorithmically, much of the recent research on SRN has focused on modeling the coupling between human motion prediction and robot control. This has motivated a class of approaches on cooperative collision avoidance (CCA) [23, 30, 32, 34, 56, 60–62, 64, 65] exploiting the observation [68] that humans tend to share the responsibility for conflict resolution in navigation domains. Many of these approaches make the cooperation assumption explicit [34, 60–62], whereas others [11, 45–47, 71] model cooperation implicitly in the way it is represented in training data. Some approaches relax this assumption, integrating asymmetric cooperation levels [31, 38, 59, 64].

Empirically, research has focused on developing methodologies for realistic validation of SRN systems. Field studies have captured informative multi-user interactions directly in deployment environments [9, 50, 63, 64], while others have used lab study paradigms [3, 26, 30, 32, 38, 43, 51], designed to elicit informative, challenging, and natural human-robot encounters. These have emphasized evaluation of control algorithms and collection of interaction data, providing valuable insights for designing SRN systems.

In this work, we build upon lab study paradigms found in prior work [31, 43] to methodically investigate the integration of HMP modeling into SRN systems deployed in constrained, simulated workspaces. Inspired by a preliminary investigation [38], our study focuses on the implications of HMP performance for robot navigation performance and human impressions. To this end, we produce a series of robot controllers based on a standard model predictive control (MPC) framework by integrating five different HMP models representing a wide range in performance. We instantiate our study at both the University of Michigan (UM) in the United States and LAAS-CNRS (LAAS) in Toulouse, France, leveraging two different robot platforms (see Fig. 1). Through a large-scale (N=80) within-subjects user study, we investigate the connection between the Average Displacement Error (ADE) — the standard measure of model performance in HMP research [1, 11, 38, 44–47, 70, 71] — and quantitative measures of robot performance and user impressions. We first conduct a confirmatory hypothesis-based analysis in which we find evidence that ADE in itself is an insufficient predictor of downstream SRN performance, and our chosen CCA-based HMP models do not improve over simpler baselines. We then conduct a deeper exploratory analysis into where these limitations come from, including findings that there remains high variability among trajectories which achieve the same ADE, CCA is often violated by users, who may act uncooperatively with co-navigating robots, and efficient robot navigation often degrades user comfort and navigation efficiency. Our study contributes critical insights for designing SRN systems that enable a mutually productive coexistence of humans and robots in realistic environments.

## 2 Related Work

**Human Motion Prediction**. The vast majority of recent approaches to HMP leverage datasets of humans navigating crowded spaces to learn prediction models. Several methods use deep learning-based generative models to learn distributions over future trajectories given histories of positions [11, 45–47]. Other learning-based approaches include that of Ziebart et al. [72], who use inverse reinforcement learning to learn a human navigation reward function, or that of Yue et al. [71], who learn the parameters of a physics inspired interaction model based on the social force model [16]. Researchers in SRN have approached HMP by developing models based on assumptions of human behavior. Several approaches leverage simple models including static [41, 62] or constant velocity (CV) [27, 38, 58, 62]; CV in particular has been demonstrated to achieve comparable performance to state-of-the-art learning-based approaches on HMP benchmarks [52]. Other approaches use CCA assumptions, including Teja S. and Alami [61], who use graph optimization with sets of constraints capturing cooperation and obstacle avoidance, and Sun et al. [60], who assume rationality among humans and use game theoretic modeling to solve for future trajectories corresponding to Nash Equilibria.

We seek to evaluate the effects of using a variety of prediction paradigms in the context of SRN. We use static and CV predictions to evaluate the efficacy of simple HMP models. We use CoHAN (Cooperative Human-Aware Navigation Planner) [62] as a representative of CCA model-based methods due to its design for use in close-proximity environments. Finally, we use the Human Scene Transformer (HST) [46] to evaluate deep learning methods due to its success on the JRDB 2023 Trajectory Prediction Challenge, which evaluated prediction embodied on robot hardware.

**Social Robot Navigation**. Over the past few decades, a wide variety of approaches leverage crowd simulation and reinforcement learning [6] to learn SRN policies, including attention [5, 27, 28] and graph-convolution [4] based architectures which model crowds as fully connected graphs. Xie and Dames [69] instead use convolutions on gridded map-representations. Others leverage SRN datasets for imitation learning, including Pokle et al. [39] who train a local controller from a simulated dataset, and Karnan et al. [19], who train a navigation policy on a dataset collected via teleoperation. An alternate line of work designs policies using techniques from modern control theory. For instance, Mavrogiannis et al. [32] design a cost function to encourage robots to commit to a passing side within a MPC architecture, whereas Samavi et al. [48] jointly tackle HMP and robot planning, satisfying safety constraints. Many MPC approaches leverage models based on CCA, including those of Sun et al. [60], Teja Singamaneni et al. [62] noted above, and Mavrogiannis and Knepper [34].

Our work builds upon the experiment of Poddar et al. [38], who demonstrate that high HMP quality does not necessarily translate to improved SRN performance. Our work seeks to provide a deeper, statistical perspective on the relationship between HMP and SRN by performing a larger-scale user study leveraging a wider set of HMP models, user impressions, two different robot embodiments, and an exploratory analysis with insights into *why* limitations of HMP performance and metrics arise.

**Benchmarking in Social Robot Navigation**. Some works evaluate performance in field studies. Thrun et al. [63] study a long-term deployment of a robot tour guide in a museum, whereas Trautman et al. [64] evaluate their algorithm in dense crowd scenarios in a university cafeteria. Satake et al. [50] perform a field study in a shopping mall to evaluate navigation behavior for robots to approach humans. Fujioka et al. [9] evaluate their intent-conveying navigation algorithm over six days in store hallway passing interactions. These works provide high ecological validity, but are expensive in time and capital requirements, motivating benchmarking performance in controllable, laboratory settings. Laboratory studies often focus on one-on-one interactions in hallways or at corners [37, 53, 54]. However, an important requirement for real-world deployment is the ability to handle multiagent interactions. To address those, Kobayashi et al. [21] contribute a reproducible experimental protocol to evaluate their method in a high-density lab setting. Landolfi et al. [26] adopt a similar workspace setup to ours, mirroring industrial settings, however the purpose of their study was to evaluate robot navigation strategy adaptation using working-memory architectures.

Our work builds upon the study of Mavrogiannis et al. [31], which explores the benefits of legibility for crowd navigation among groups of three people. Our experiment design gives rise to dense interactions between a robot and two people (simplifying subject recruiting), allowing for uninterrupted, natural human-robot encounters in a more constrained workspace, under a simpler distraction task. Our experiment design is further motivated recent experimental insights [58] suggesting that to see real-world progress in SRN, it is crucial to scale testing in settings of high complexity, motivating us to design for dense, close interactions.

## 3 Problem Setting

We formalize the problem of SRN and describe a general model predictive control architecture for tackling it.

**Social Robot Navigation**. We consider a robot navigating among $n \geq 1$ humans in a workspace $\mathcal{W} \subseteq \mathbb{R}^2$. We describe the state of the robot as $s^r \in \mathcal{W}$ and the state of human $i \in \{1, \ldots, n\}$ as $s^i$. The state of the robot evolves according to dynamics $s^r_{t+1} = g(s^r_t, u_t)$, where $u_t$ is a control action (speed, and steering angle), drawn from a space of controls $\mathcal{U}$. The robot starts from an initial configuration $s^r_0$ and moves towards a goal $d^r$ by following a policy $\pi^r$, while humans navigate from their initial configurations $s^i_0$ towards their goals, $d^i$ by following a policy $\pi^i, i \in \mathcal{N}$; agents' goals and policies are unknown to one another. We assume that the robot perfectly observes human states at all times. We focus on the problem of designing a robot policy $\pi^r$, such that the robot reaches its goal $d^r$ in a safe, efficient, and socially compliant fashion.

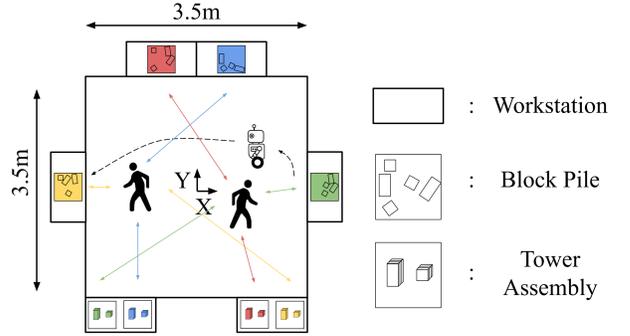

**Figure 2: Experimental setup. Each participant is responsible for assembling towers from two different colored (green and red or yellow and blue) piles of blocks, while the robot traverses the workspace between stations. The color ordering is fixed so the green and yellow towers are completed before the red and blue.**

**Social Robot Navigation as Model Predictive Control (MPC).** A general way [13, 38, 48, 49, 58] to determine $\pi_R$ is to formulate social navigation as a receding-horizon optimal control problem, optimized over a horizon $T$:

$$u^*_{t:T} = \arg\min_{u_t \in \mathcal{U}} \sum_t^{T-1} \mathcal{J}(s^r_{t+1}, s^{1:n}_{t+1})$$
$$s.t. \ s^r_{t+1} = g(s^r_t, u_t) \qquad (1)$$
$$s^{1:n}_{t+1} = f(s^r_{t-h:t}, s^{1:n}_{t-h:t}),$$

where $\mathcal{J}$ is a cost function capturing qualities of SRN, and $f$ is an HMP model that takes as input the state history of the robot $s^r_{t-h:t}$ and humans $s^{1:n}_{t-h:t}$ over a window $h$, and outputs a prediction of human motion, $s^{1:n}_{t+1}$. Following prior work [32, 38], we model the cost to capture goal-directedness, social awareness, and obstacle avoidance – all fundamental objectives for a robot completing tasks in human environments. Specifically, we model cost $\mathcal{J}$ as:

$$\mathcal{J}(s^r, s^{1:n}) = a_g \mathcal{J}_g(s^r) + a_d \mathcal{J}_d(s^r, s^{1:n}) + a_o \mathcal{J}_o(s^r), \quad (2)$$

where $\mathcal{J}_g(s^r) = (s^r - d^r)^2$ penalizes distance to the robot's goal, $\mathcal{J}_d(s^r, s^{1:n}) = \sum_{i=1}^n A_d^2(s^r, s^i)$, penalizes violations to human personal space as done in prior work [20], and $\mathcal{J}_o(s^r) = \mathbb{1}(A_R(s^r) \cap \mathcal{W}^c \neq \emptyset)$ penalizes violations of the workspace boundary, where $A^r(s^r)$ is the volume occupied by the robot at pose $s^r$. Weights $a_g$, $a_d$, and $a_o$ represent the relative importance of each cost.

## 4 User Study

We detail the experiment design for our user study (IRB HUM0259961 and Université Toulouse Ethics Approval 2024_917).

### 4.1 Scenario Design

Studying group interactions with a navigating robot in a lab environment is challenging: we want to ensure natural human motion, while motivating nontrivial human-robot interactions that allow us to stress-test the SRN system. Our design builds upon prior work designing SRN scenarios in a lab setting [26, 30], and the constrained, goal-directed design mirrors many domains of interest for SRN, including logistics, manufacturing, and healthcare.

**Subject Task.** Subjects were instructed to complete a series of tower assembly tasks on six small tables (hereafter referred to as workstations) placed along the perimeter of the workspace. To ensure subjects consistently crossed paths with each other and the robot, they were each assigned to assemble towers from two colors of blocks from piles on opposite sides of the workspace (i.e. red and green or yellow and blue in Fig. 2). Blocks of each color were already placed at the bottom two workstations (two colors at each station) to serve as the bases for constructing towers from the piled blocks. To further encourage interaction and to decrease variance in strategy, participants were also instructed to complete their colors in order from close to far from the bottom of the workspace (i.e. yellow and green, then blue and red). To ensure subjects did not rush or strategize with each other to simplify their interactions, they were instructed to walk at a natural pace and not to talk.

**Robot Task.** Subjects were told the robot's task was monitoring their progress, which it completed by visiting workstations (goals). The goal sequence was generated randomly subject to the y-axis always being crossed, which resulted in regular crossings with both users. To reduce learning effects, the sequence was flipped across the y-axis each trial. This design is motivated by the vision of enabling robots and humans to work in a constrained shared space on practical tasks commonly encountered in fulfillment and manufacturing, such as assembly and sorting.

## 4.2 Experimental Procedure

At each study session, two subjects completed five trials, each involving navigation alongside the robot under a different condition. Each condition corresponded to the robot integrating a different HMP model $f$ into the same MPC architecture from eq. (1). Before the study, subjects were asked to read and sign a consent form and offered the option to opt out of video recording and data publishing. Subjects were informed the study was about user impressions of robot navigation algorithms, but not about the specific purpose of the study. After obtaining consent, subjects were briefed on the components of the workspace, their task, and the robot's role. Subjects then completed a practice trial in which the robot remained static at its starting position. They then completed five trials, filling out a survey on lab computers right after each trial, while a study team member reset the workspace. To account for ordering effects, the condition order was determined using a balanced Latin square. Upon completing all five trials and surveys, subjects completed demographic surveys, were given $20 in compensation, and debriefed with an information sheet describing the purpose of the study. The total study time was a maximum of 50', with 15' spent on forms, instructions, and practice, 30' on trials, and 5' on debriefing.

**Trial Description.** Subjects started and ended each trial at 'X' shapes marked on the ground in front of the two assembly stations, while the robot began in the center of the workspace facing the subjects. A study team member asked subjects for verbal confirmation that both were ready to begin the trial. Once confirmation was received, the robot would start, and the trial would begin. The trial concluded when both subjects completed their towers and returned to their starting positions. Trials took a mean of 204.5s.

## 4.3 Conditions

We followed a within-subjects design, so all subjects experienced the same five conditions, each corresponding to navigation alongside a robot integrating a different HMP model within the MPC. In early pilots, we found that five was the maximum number of conditions before participants became fatigued. While numerous HMP models have been proposed, we made a selection balancing a wide range of model performance and representation in prior work:

**No Prediction (NP).** The robot does not perceive the humans navigating around it. NP is representative of the common practice in industrial robotics of unreactive robots [14, 15], appears as a baseline in physical SRN user studies [26, 35], and has been shown to perform reasonably effectively in simulated SRN studies [33, 58].

**Static (ST).** Humans are predicted to stay in place over the full time horizon. ST represents the simplest human motion model which still allows for human aware planning and control, and typically appears in costmap-based approaches [41, 62].

**Constant Velocity (CV).** Humans are treated as dynamic obstacles with no additional modeling. CV has repeatedly been shown to be an effective model of human navigation in terms of prediction accuracy [52] and for downstream use in SRN [22, 27, 38, 58, 62].

**Human Scene Transformer (HST).** HST [46] is a transformer-based generative model which outputs a Gaussian mixture model distribution of future joint trajectory modes using a joint trajectory history. To obtain the final prediction, we sample a single joint trajectory from the most likely mode, and model the influence of the robot by including it as an agent in the joint trajectory prediction.

**CoHAN.** CoHAN [56, 61, 62] is a CCA trajectory planner that jointly optimizes human-robot trajectories. It optimizes the robot trajectory using social costs including human-human and human-robot safety, human visibility and speed, and obstacle avoidance.

## 4.4 Measures

**Robot performance**. We measure navigation performance with respect to the robot's own motion using objective metrics of productivity, path directness, and motion smoothness. Robot productivity is measured in Goals Per Second (GPS, $goals/s$), while we leverage the Robot Path Irregularity (PI, $rad/m$), which is the amount of unnecessary rotation (angle between the agent heading and their goal) per unit of path length, and Robot Average Acceleration (AA, $m/s^2$), which are both used in existing social navigation literature [10], to capture path directness and sudden speed changes, respectively.

**Human performance**. We measure navigation performance with respect to aspects of motion and impressions induced in humans. We use the same objective measures as for robot motion, that is Human GPS, Human PI, and Human AA, again following prior SRN studies [30]. Higher Human PI and Human AA represent deviations from direct, smooth motion which humans typically achieve when maximally productive and comfortable. We also collect subjective measures of Discomfort using the Robotic Social Attributes Survey (RoSAS) Discomfort questions, rated on a 9-point Likert scale. Our subjective measures of subject Perceived Workload are collected using the using the NASA TLX survey, which has 6 types of demand (*Mental, Physical, Temporal, Frustration, Performance, Effort*) rated on 21-point Likert scales. In addition to the above subjective and objective metrics, we also collected open-ended impressions from users via a survey prompt to "leave any additional comments" at the end of each individual algorithm survey.

**Team performance.** In addition to robot and subject productivity separately, we capture overall team performance with *Team GPS*, computed per-subject by averaging Human and Robot GPS.

### 4.5 Hypotheses

While the importance of understanding the relationship between HMP model quality and SRN performance has been pointed out in prior work [38], we are not aware of any prior studies making it precise (e.g., statistical evaluation, analysis of user impressions, diversity of methods). To this end, we ask the question: *how well does HMP performance correlate with SRN performance?* To investigate this question, we use the Average Displacement Error (ADE) as a measure of model performance, where $ADE = \frac{1}{T}\sum_{t=1}^{T}\|s_t - s_t^*\|$, $s_{1:T}$ is a predicted trajectory and $s_{1:T}^*$ is the ground truth. ADE is the primary metric adopted by the HMP community [44] for assessing long-horizon model performance, thus it is essential for understanding the connection between HMP model performance and downstream SRN performance. Additionally, we expect that CCA-based models (i.e., HST and CoHAN) will outperform models treating humans as obstacles or ignoring them entirely (NP, ST, and CV). We codify these expectations into the following hypotheses:

**H1:** *Lower HMP error results in:*

1a) *Higher Productivity*, measured in terms of Team GPS.
1b) *Lower Workload*, measured in terms of user impressions (NASA TLX) and human path directness (Human PI).
1c) *Lower Discomfort* (RoSAS) and human path smoothness (Human AA).

**H2:** *CCA-based HMP results in:*

2a) *Higher Productivity*, measured in terms of Team GPS.
2b) *Lower Workload*, measured in terms of user impressions (NASA TLX), and human path directness (Human PI).
2c) *Lower Discomfort* (RoSAS) and human path smoothness (Human AA).

### 4.6 Implementation Details

**Site details.** We used a square-sized (side 3.5m) workspace covered by motion capture across both Sites. This size ensured that users and robot were constantly inside each others' social proxemic zones [12], and were regularly made to enter each others' personal zones [12], representing very close interactions. Motion capture fed realtime poses of subjects and the robot to all algorithms. The procedure and data collection were implemented identically across both sites with the exception of the robot platform: Hello Robot Stretch ($330 \times 340 \times 1410$ mm, 24.5 kg) was used at UM; Willow Garage PR2 ($668 \times 668 \times 1330$ mm, 226.8 kg) was used at LAAS (see Fig. 1). Mirroring the study setup allowed us to increase the study size and population diversity, as well as increase the applicability of our findings to distinct robot morphologies. We have accounted for the site difference in our statistical analysis in Sec. 5.

**HMP.** To ensure fair comparison, all HMP models operated using $0.4s$ timesteps, with $3.2s$ (8 timestep) histories and $4.8s$ (12 timestep) prediction horizons, matching standard evaluation settings in the HMP literature [11, 45–47], and used the same controller parameters. The HST checkpoint produces $m = 20$ prediction modes, and was trained on the ZARA1 portion of the ETH/UCY dataset, which has been shown to provide a good basis for learning multi-pedestrian interactions [8]. In pilot experiments, we found participants rated the robot more favorably using a ZARA1 trained HST compared to a JRDB [29] trained HST. To match other models, CoHAN was allowed to plan for agents outside the robot FoV, and had no backoff behavior. To prevent the predictions from HST, CV, and CoHAN from crossing the workspace edges, we truncated predictions at the edges by replacing all out-of-bounds predicted states with the latest state of the prediction still inside.

**MPC.** Our MPC architecture is based on a GPU-parallelized implementation of the Model Predictive Path Integral (MPPI) [25, 67] using 2000 trajectory samples. The control architecture and prediction models were identical at both sites. The maximum speed of the robot was capped at 0.3m/s, as it was found to be a safe limit for our setup. Differential drive kinematics were adopted at both sites for consistency. Upon reaching a goal, the robot would turn in place towards its next goal before continuing to navigate. The full prediction and control loop closed at 50 Hz for NP, ST, and CV, 20 Hz for HST, and 8 Hz for CoHAN due to differences in computational burden for each model. Early pilots at both sites verified that these frequencies were sufficient to ensure reactivity under our experimental settings. HMP models and MPC were verified and tuned on hardware at both sites to ensure consistency.

## 5 Analysis

A total of 96 subjects participated in 48 sessions across both sites. One trial was aborted at UM due to technical difficulties, and seven trials (six at UM and one at LAAS) were not included in the analysis due to technical errors unknown during the sessions. The final dataset includes 80 subjects from 40 sessions (20 sessions at each site). UM experiments involved 40 subjects (11 female, 28 male, 1 unidentified), with robotics experience $3.48/5.00 \pm 1.38$, and ages: 18-22 (20); 23-29 (20). LAAS involved 40 subjects (12 female, 28 male), with robotics experience $3.64/5.00 \pm 1.04$, and ages: 18-22 (1); 23-29 (28); aged 30-39 (11). Gender and robotics experience were not significantly different between sites; UM was younger ($p < 0.001$). We model statistical relationships between pairs of metrics in Sec. 4 using linear mixed-effects regression models, with the predictor metric as a fixed effect, *site* as an interaction term on the metric (capturing differences between sites), and random effects for individual participants and experiment groups. Metrics are calculated per trial as averages, with timesteps in which the robot was turning to its next goal removed. For significance testing, we perform Likelihood Ratio Tests between the full model and nested models lacking the predictor metric fixed effect, or site interaction. Pairwise comparison is implemented using Tukey's Honestly Significantly Different (HSD) Test; **H1** and the exploratory analysis p-values are corrected using the Holm method.

**Table 1: Estimated Marginal Mean ADE by prediction model. The top values are from UM, while the bottom are from LAAS.**

| Algorithm | Mean ADE (m) | Lower CL | Upper CL |
|-----------|--------------|----------|----------|
| ST | 1.44 | 1.38 | 1.51 |
| CV | 1.15 | 1.09 | 1.22 |
| HST | 1.20 | 1.15 | 1.27 |
| CoHAN | 1.51 | 1.46 | 1.58 |
| ST | 1.25 | 1.19 | 1.31 |
| CV | 0.93 | 0.87 | 0.99 |
| HST | 1.09 | 1.04 | 1.16 |
| CoHAN | 0.95 | 0.90 | 1.02 |

**Table 2: Effects of ADE on metrics.**

| Metric | UM Est. | LAAS Est. | Pr(>Chisq) |
|--------|---------|-----------|------------|
| Discomfort | -0.2135 | -0.9060 | 0.1544 |
| **Team GPS** | 0.0138 | 0.0315 | <0.001*** |
| **Human GPS** | 0.0498 | 0.0361 | <0.001*** |
| **Robot GPS** | -0.0249 | 0.0137 | <0.001*** |
| **Human AA** | 0.2507 | 0.12214 | <0.001*** |
| **Human PI** | 11.467 | -13.663 | <0.001*** |
| Frustration | -0.3072 | -2.9251 | 0.0556 |
| Mental | -0.2035 | 0.7716 | 0.6543 |
| Physical | -0.3557 | -0.9487 | 0.3694 |
| Temporal | -1.720 | -0.6470 | 0.2268 |
| Effort | 0.7776 | -1.7077 | 0.1307 |
| Performance | 1.5376 | 0.6988 | 0.5781 |

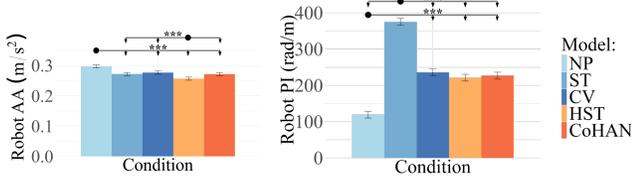

**Figure 3: Estimated Marginal Means by model for robot motion metrics. Relationships between one model (•) and others (↓) have significance levels denoted $p < 0.05^*$, $p < 0.01^{**}$, and $p < 0.001^{***}$.**

### 5.1 How Prediction Error shapes Robot Navigation Behavior

We first verify that our HMP model set spans a wide range of performance. Using Wald $t$-tests we validate that all pairwise differences in the ADE of different models are significant ($p < 0.01$), confirming that all models have distinct error levels (Table 1). We next validate that, beyond prediction error, each HMP model also results in distinct robot navigation behavior. NP exhibits lower PI but higher GPS compared to other methods ($p < 0.001$), whereas ST exhibits higher PI but lower GPS ($p < 0.05$) (see Fig. 3, Fig. 4): NP acts efficiently as an essentially blind planner whereas ST acts conservatively due to poor prediction (see Table 1). HST's AA is lower than CoHAN and CV ($p < 0.001$, see Fig. 3), a sentiment echoed by a comment that the HST "drive was a a lot smoother compared to the other ones." Qualitatively, users found CoHAN "less smart," whereas CV received comments describing it as "aware."

### 5.2 How Prediction Error shapes Productivity and Impressions

We list significance relationships between ADE and collected metrics in Table 2. We find that lower ADE is correlated with *lower* Team and Human GPS ($p < 0.001$) at both sites, indicating that human and team productivity did not increase (see Fig. 4) as prediction error decreased. Thus we find that **H1a** is not supported. Regarding workload, we find that lower ADE is correlated to lower Human PI at UM, but higher at LAAS, whereas no significant relationships were found between ADE and NASA TLX, suggesting that humans did not take more direct paths and perceive lower workload, suggesting that **H1b** was not supported. Finally, we did not find a significant relationship between ADE and Discomfort (RoSAS), but did find that lower ADE is correlated with *lower* Human AA ($p < 0.001$), suggesting that subjects moved more smoothly when prediction error dropped but did not perceive themselves as more comfortable. Thus, we conclude that **H1c** was partially supported: lower ADE error *does not* directly correlate to improved productivity and user impressions, suggesting **H1** is not supported.

### 5.3 How CCA shapes Productivity and Impressions

We perform a comparative analysis of CCA (HST and CoHAN) against other models via pairwise comparisons (see Fig. 6). In terms of Team GPS, we find that CoHAN outperformed NP ($p < 0.01$), but both CoHAN and HST were outperformed by NP ($p < 0.001$). Thus, overall, we find that our CCA models did not lead to definitive team productivity improvement, indicating that **H2a** is unsupported. We find that our CCA models did not decrease subject workload: CoHAN was worse than NP for Frustration ($p < 0.05$) and saw higher Human PI than all other methods ($p < 0.001$); HST saw no significant pairwise differences. Thus we find that subjects took less direct paths and experienced higher workload around CCA models, so **H2b** is unsupported. HST exhibited *lower* perceived Discomfort than NP ($p < 0.01$), but similar to CV and ST. CoHAN had *higher* Discomfort than CV and ST (all $p < 0.05$) and similar Discomfort to NP. In terms of navigation smoothness, we find that HST saw lower Human AA than CV and ST ($p < 0.001$). This was also reflected in it receiving the most positive comments, with users describing it as "less near," "more aware," or "[...] remote controlled by a human." However, NP saw the lowest Human AA ($p < 0.001$), with comments that it "seemed more predictable." These findings suggest our chosen CCA methods had mixed comfort perception and subject navigation smoothness effects, and thus support for **H2c** is mixed. Overall, we find that **H2** is not supported.

### 5.4 Exploring the Limitations of ADE for Characterizing Prediction Performance

We provide additional exploratory analysis into *why* ADE failed to correlate with navigation performance. Our insight is that the failure is due to high variability in predictions that achieve the same ADE. This variability results in predictions with the same ADE exhibiting vastly different alternative characteristics. This is because ADE represents a first order error over the predicted position, and therefore it misses higher-order properties of motion, which may be crucial to understand for SRN. For instance, a prediction which is erroneous due to underestimating the human's velocity may cause the robot to incorrectly believe it has time to pass in front of the human, causing it to block their path. We confirm this variability among predictions with similar ADEs by binning predictions in intervals of 0.05 ADE and calculating two additional measures of prediction error: *Velocity Difference* and *Path Irregularity Difference*, which capture respectively the difference between the ground-truth and predicted speed and directness, averaged over all timesteps in the same manner as ADE. As shown in Fig. 5, even for low ADE, the variance of both metrics remains high. In principle, it is expected that variation across Velocity and Path Irregularity Differences would drop as ADE approaches 0. However, at current model performance levels in challenging settings [71], we see that ADE is insufficient for identifying the best performing predictions.

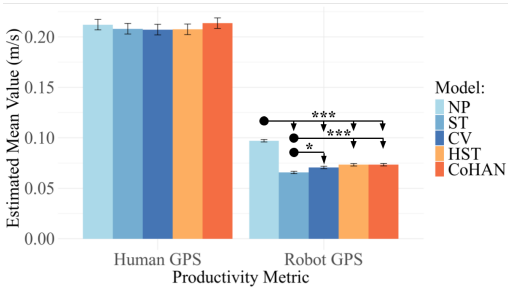

Figure 4: Productivity metric estimated means by model. Relationships between one model (•) and others (↓) have significance levels denoted $p < 0.05^*$, $p < 0.01^{**}$, and $p < 0.001^{***}$.

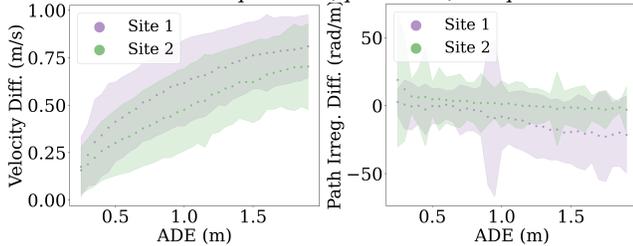

Figure 5: Insufficiency of average displacement error (ADE) as a measure of accuracy. Dotted lines indicate means and error bands indicate standard deviations. We see that for each ADE level, the velocity and PI differences exhibit high variance.

## 5.5 Exploring the Effects of CCA Violations on Prediction and Navigation Performance

We provide additional exploratory analysis into *why* CCA models saw no or limited performance improvements over other models. CCA models generally assume that humans will act cooperatively. Some models also assume that the cooperation between all parties will be equal [60, 62, 65]. Our insight is that violation of the CCA assumption may lead to model performance drops, thus we seek to estimate the cooperativeness of the subjects in our study. As a proxy for assessing cooperativeness, we use the metric of *Responsibility* [40, 66], which captures the observed share of each agent in resolving future collisions. A positive *Responsibility* from one agent to another means that the agent is moving to avoid a future collision, and thus acting cooperatively towards the other agent. A value of 0 indicates that the two agents either will not collide if they maintain their current trajectories, or that the agent is acting neutrally by not changing its trajectory. A negative *Responsibility* indicates that an agent is taking actions that move its future trajectory towards a more direct future collision.

Table 3: Average Responsibility for collision avoidance for humans and the robot. Higher values indicate more cooperative navigation. H refers to Human and R to Robot.

| Site | RH Resp. | HR Resp. | Diff. | CI (Low) | p-value |
|------|----------|----------|-------|----------|---------|
| UM   | 0.0316   | -0.0412  | 0.0728 | 0.0665  | < 0.001*** |
| LAAS | 0.1830   | 0.0240   | 0.1590 | 0.1290  | < 0.001*** |

We perform paired one-sided t-tests to compare Responsibility *from subjects to the robot* (HR) and *from the robot to subjects* (RH). We find that RH Responsibility was significantly higher than HR at both sites ($p < 0.001$), and that HR Responsibility was significantly higher at LAAS ($p < 0.001$) than HR at UM (Table 3), indicating subjects were not equally reciprocating the robot's cooperative behavior, and that at UM they were, on average, not even minimally cooperative. CoHAN relies not only on cooperation, but *symmetric* cooperation between humans and the robot. Thus we believe this reduced its effectiveness, particularly at UM, as we note the gap in ADE between sites for CoHAN is larger than for other models (see Table 1). We believe that HST was more robust due to its ability to model uncooperative and asymmetrically cooperative behavior in its training data. We speculate the difference in cooperativeness per site is influenced by robot morphology: the smaller Stretch could more easily be ignored, in contrast to the massive PR2.

## 5.6 Exploring the Coupling of Human and Robot Motion Efficiency

The connection between HMP models and human motion and impressions is indirect: subjects experienced different HMP models via their resulting robot motion. Here, we provide additional insight into the direct, coupled relationship between robot and human motion– a connection long discussed in the literature [18, 24, 36, 42]. To explore this coupling, we analyze the correlations between our metrics of robot and human motion from Sec. 4, with the addition of *Average Speed (AS)*, which represents navigation efficiency. We find that when the robot took longer paths (higher Robot PI), humans were able to move faster (higher Human AS) at both sites ($p < 0.01$). Furthermore, when the robot moved faster (higher Robot AS), humans took less direct paths (higher Human PI at LAAS ($p < 0.001$). We also see that Human and Robot PI are inversely correlated at LAAS ($p < 0.001$), meaning that humans take more direct paths when the robot takes less direct paths, and vice versa. Finally, we see that faster, more direct robot motion (higher Robot AS and lower Robot PI) correlated with higher Discomfort at LAAS ($p < 0.05$). These observations suggest deploying robots in crowded, dynamic spaces presents a design challenge: efficient robot motion can degrade human efficiency and comfort; conversely, unilaterally prioritizing humans makes robot motion inefficient.

## 5.7 Exploring the Effects of Site on Subjects

Users had significantly different impressions and navigation behavior by site. Compared to LAAS, UM saw smoother (lower Human AA; $p < 0.001$) and more direct human motion (lower Human PI; $p < 0.001$), and higher productivity (Human GPS; $p < 0.001$).

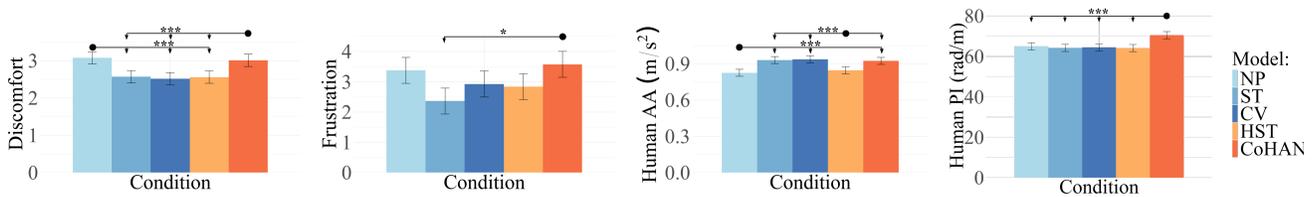

Figure 6: Estimated Marginal Means by model for subjective and objective user metrics. Relationships between one model (•) and others (↓) have significance levels denoted $p < 0.05^*$, $p < 0.01^{**}$, and $p < 0.001^{***}$.

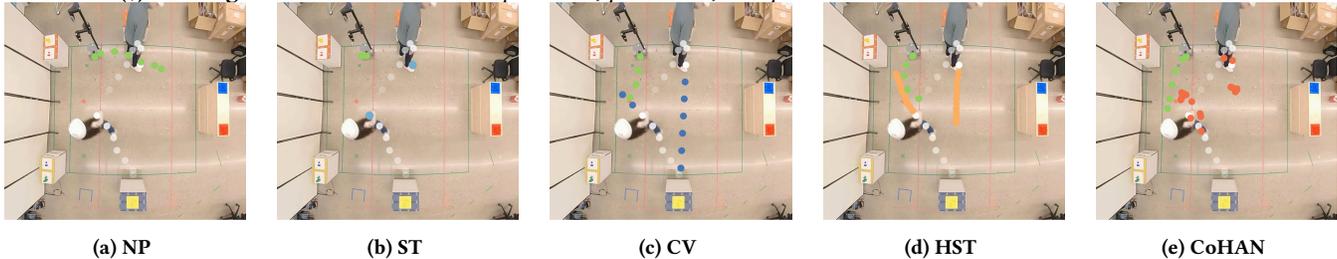

(a) NP      (b) ST      (c) CV      (d) HST      (e) CoHAN

Figure 7: Examples of predictions and their distinctive effects on resulting robot navigation. Predictions are in blue or orange, trajectory history used for prediction is in white, and the minimum cost MPPI rollout based on the predictions is in green.

However, users at UM expressed significantly *higher* Discomfort than at LAAS ($p < 0.05$). We speculate that these differences arise in part due to robot morphology differences, although we note that we cannot isolate morphological effects from other factors that varied by site. The smaller robot at UM may have allowed users to more easily maneuver around it, allowing for faster, smoother human motion. While at LAAS the larger robot may have led users to yield more often to it when blocking their goal, at UM, users more often ignored the smaller robot and reached around it to pick or place their block. This led to a UM phenomenon in which the robot would become "stuck" near goals, causing repeated, uncomfortable interactions by blocking a user's goal (see video supplement).

## 6 Discussion

**Insufficiency of ADE**. Lower ADE did not correlate with improved navigation performance and user impressions. The challenges of HMP, including human stochasticity and variability, will always introduce errors into prediction; these errors may arise differently for each HMP model, even when their ADE is equal. This motivates finding additional informative ways to characterize HMP performance for SRN. This will be critical for effectively leveraging HMP for proactive maintenance of acceptable proxemic zones [12], cooperative space negotiation [35, 36], and safe decision making [48, 49] on mobile robots deployed in human environments.

**Humans are not as cooperative as previously thought**. We showed that humans took significantly less responsibility for collision avoidance with the robot than the robot did with them. Our findings validate prior CCA models that account for asymmetric cooperation [59, 64], and are in line with empirical work showing that humans often apply different social norms to robots [42]. Additionally, our finding that cooperation was lower at the site with the smaller robot suggests morphology may be an important yet overlooked factor impacting CCA modeling.

**Discomfort and efficiency tradeoffs**. Our exploratory analysis identified that efficient robot navigation may compromise human efficiency and comfort. Prior work has investigated the tradeoff between human comfort and workload and robot efficiency [2, 57].

Our findings that HST induced smoother human paths than CV and ST while maintaining team productivity demonstrate that CCA models can balance human cognizance with robot productivity.

**Limitations**. While our scenario motivates realistic human-robot spatial interactions, it is still an abstraction of those encountered in real workspaces. We integrated HMP models without additional adaptation to ensure our insights are directly relevant to the HMP literature [1, 11, 46, 47], and because tailoring HMP models to constrained settings is an open problem. HMP is a rich area of research, and more work is needed to evaluate the effectiveness of other recent models. Our MPC followed prior work to optimize fundamental SRN objectives [32, 33, 38], but alternative objectives [17] may lead to different interactions. While our suite of metrics is ubiquitous in SRN studies [10, 31, 33], their connection with user impressions is part of ongoing investigation. Future work may provide additional insight into the links between motion characteristics and user impressions. Finally, site differences likely arose through a combination of factors including robot morphology, culture, and demographics. While several of these are consistent with the larger size of the PR2 being perceived differently than the smaller Stretch, we cannot disentangle any morphological effects from other site differences. Similarly, while part of the site differences might be attributed to population differences, we cannot validate that.

## Acknowledgements

This work was partially supported by the Horizon Europe Framework Programme grant 101070596 - euROBIN. The authors thank Mo Xu for his contributions to the HST model implementation.